\documentclass[10pt, a4paper]{article}
\usepackage{lrec}
\usepackage{multibib}
\newcites{languageresource}{Language Resources}
\usepackage{graphicx}
\usepackage{tabularx}
\usepackage{soul}

\usepackage{epstopdf}
\usepackage[latin1]{inputenc}
 
\usepackage{hyperref}
\usepackage{xstring}

\title{ \textbf{WASA: A Web Application for Sequence Annotation}}

\name{Fahad AlGhamdi, and Mona Diab}

\address{Department of Computer Science \\
         The George Washington University \\
         Washington, DC\\
         \{fghamdi, mtdiab\}@gwu.edu\\}

\abstract{
Data annotation is an important and necessary task for all NLP applications. Designing and implementing a web-based application that enables many annotators to annotate and enter their input into one central database is not a trivial task. These kinds of web-based applications require a consistent and robust backup for the underlying database and support to enhance the efficiency and speed of the annotation. Also, they need to ensure that the annotations are stored with a minimal amount of redundancy in order to take advantage of the available resources(e.g, storage space). In this paper, we introduce WASA,  a web-based annotation system for managing large-scale multilingual Code Switching (CS) data annotation. Although WASA has the ability to perform the annotation for any token sequence with arbitrary tag sets, we will focus on how WASA is used for CS annotation.  The system supports concurrent annotation, handles multiple encodings, allows for several levels of management control,  and enables quality control measures while seamlessly reporting annotation statistics from various perspectives and at different levels of granularity. Moreover, the system is integrated with a robust language specific date prepossessing tool to enhance the speed and efficiency of the annotation. We describe the annotation and the administration interfaces as well as the backend engine.\newline \newline  }

\begin{document}

\maketitleabstract

\section{Introduction}

Code Switching (CS) is a phenomenon that occurs when multilingual speakers alternate between more than one language or dialect. This phenomenon can be observed in different linguistic levels of representation for different language pairs: phonological, morphological, lexical, syntactic, semantic, and discourse/pragmatics. CS presents serious challenges for language technologies, including parsing, Machine Translation (MT), Information Retrieval (IR) and others. A major barrier to research on CS has been the lack of large multilingual, multi-genre CS-annotated corpora. Creating such corpora involves managing many annotators working on multiple tasks at different times, consistent and robust backups of the underlying database, quality control, etc. In this paper, we present our effort in building an annotation system, WASA, that can manage and facilitate large-scale CS data annotation.
WASA differs from other annotation systems in several respects. Our system has an option that can provide initial automatic tagging for specific tokens such as Latin words, URL, punctuation, digits, diacritics, emoticons, and speech effect tokens. This option increases the quality and the speed of annotation substantially.  Moreover, the system is integrated with language-specific date preprocessing tool Smart Preprocessing (Quasi) Language Independent Tool (SPLIT) \cite{AlBadrashiny2016SPLITSP} to streamline raw data cleaning and preparation.  

	The remainder of this paper is organized as follows: Section 2 provides an overview of related work. Section 3 describes the System Architecture. Types of users including permissions and users tasks are introduced in Section 4. The data preprocessing and cleaning are discussed in Section 5. We provide an overview of the database design in Section 6. Inter-annotator agreement, current status, and our conclusion and future work are discussed in sections 7,8 and 9, respectively. 

\section{Related Works}

	Although, many annotation tools, such as \cite{aziz2012pet}, \cite{cunningham}, \cite{kahan}, MnM \cite{vargas}, GATE (\cite{cunningham}; \cite{aswani}, and \cite{dickinson}), are effective in serving their intended purposes, none of them meets the CS annotation requirements perfectly. We need a tool that can help in sequence annotating in a way that can report the time needed for annotators to get their tasks done, manage number of annotator teams, enable quality control measures and annotation statistics, and assign some initial tags to some tags automatically (e.g. punctuation, URL, emoticon, etc.) 
     	
		
Our tool is most similar to the annotation tool for the COLABA project \cite{diab2010colaba}; \cite{benajiba2010web}(Benajiba and Diab, 2010; Diab et al., 2010). We specifically emulate the annotator management component in the COLABA annotation tool.  Although, the code switching annotation task and manual diacritization of Standard Arabic text task are completely different tasks, the MANDIAC tool \cite{obeid}, which used for diacritization annotation task, has a similar annotator management component to the WASA management component. However, the technologies used in both management components are different. For instance, WASA uses PostgreSql database to store content, while MANDIAC uses a JSON blob to store content. Two other comparable tools to ours are WebANNO \cite{WebAnnot} and SWAT \cite{samih2016sawt}. They both use the latest available technologies to perform a number of linguistic annotation types. The SWAT tool is a web-based interface for annotating tokens in a sequence with a predefined set of labels. The main advantages of this tool are the simplicity of its use and installation as it only requires a modern web browser and minimum server-side requirements to get the tool work. The WebANNO tool is also a web-based tool that offers wide range of linguistic annotations tasks, e.g., named entity, dependency parsing, co-reference chain identification, and part-of speech annotation.   

However, both systems SWAT and WebANNO lack of some functionalities and features that can simplify and speed up the annotation task for our purposes. In the SWAT system for example, there is no support for user roles. Therefore, some tasks such as managing the number of annotators, monitoring the progress of the annotators, assigning tasks given to the annotators, and ensuring the quality of the submitted annotation are difficult to handle or manage with only one user type. Moreover, both systems do not have the option that can provide initial automatic tagging for named entities (NE), Latin words, URL, punctuation, number, diacritics, emoticon, and speech effects tokens. We noticed that tagging these tokens automatically increases the speed of the annotation substantially. Finally, unlike both systems, our system can seamlessly integrate with language specific data preprocessing tool to streamline raw data cleaning and preparation. 

\section{System Architecture}

\begin{figure}
  \centering
    \includegraphics[width=0.5\textwidth]{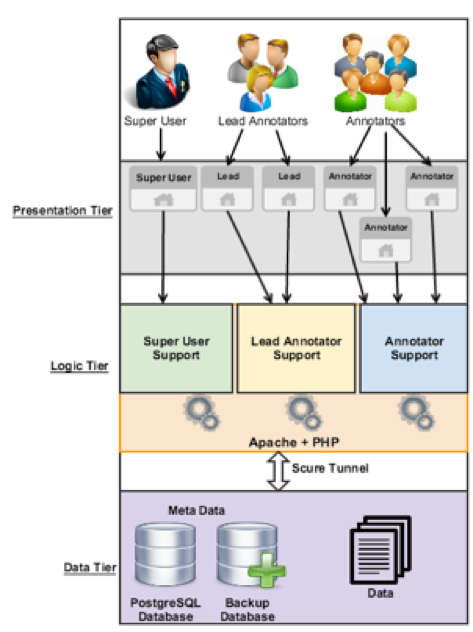}
  \caption{System Architecture}
  \label{fig:Architecture}
\end{figure}

WASA is a typical three-tier web-based application. The platform is divided into three tiers, each with a specific function. The first tier is a data tier that saves all metadata in PostgreSql database in addition to both the annotated and raw data files. All this data is stored on a file server. The second tier is a logical tier. It contains PHP scripts that interact with an Apache web server. It is responsible for all functionalities provided by the system to the different types of users. All requests are sent by the web server to the  PostgreSql database server through a secured tunnel. The third and last tier is the presentation tier. It is browser independent, which enables accessing the system from many different clients. It provides an intuitive GUI tailored to each user type. This architecture design allows multiple annotators to work on various tasks simultaneously. On the other hand, WASA allows the admin user to manage and handle a single central database. The system can handle multiple encodings allowing for multilingual processing. Figure-\ref{fig:Architecture} gives a high level overview of the tool's architecture.

\begin{figure*}[t!]
  \centering
    \includegraphics[width=1\textwidth]{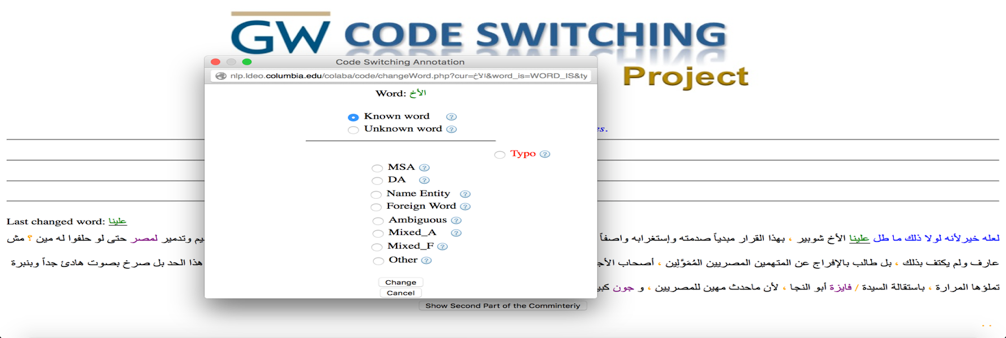}
  \caption{Annotation Screen}
  \label{fig:screen-shot-1}
\end{figure*}

\begin{figure*}
  \centering
    \includegraphics[width=1\textwidth]{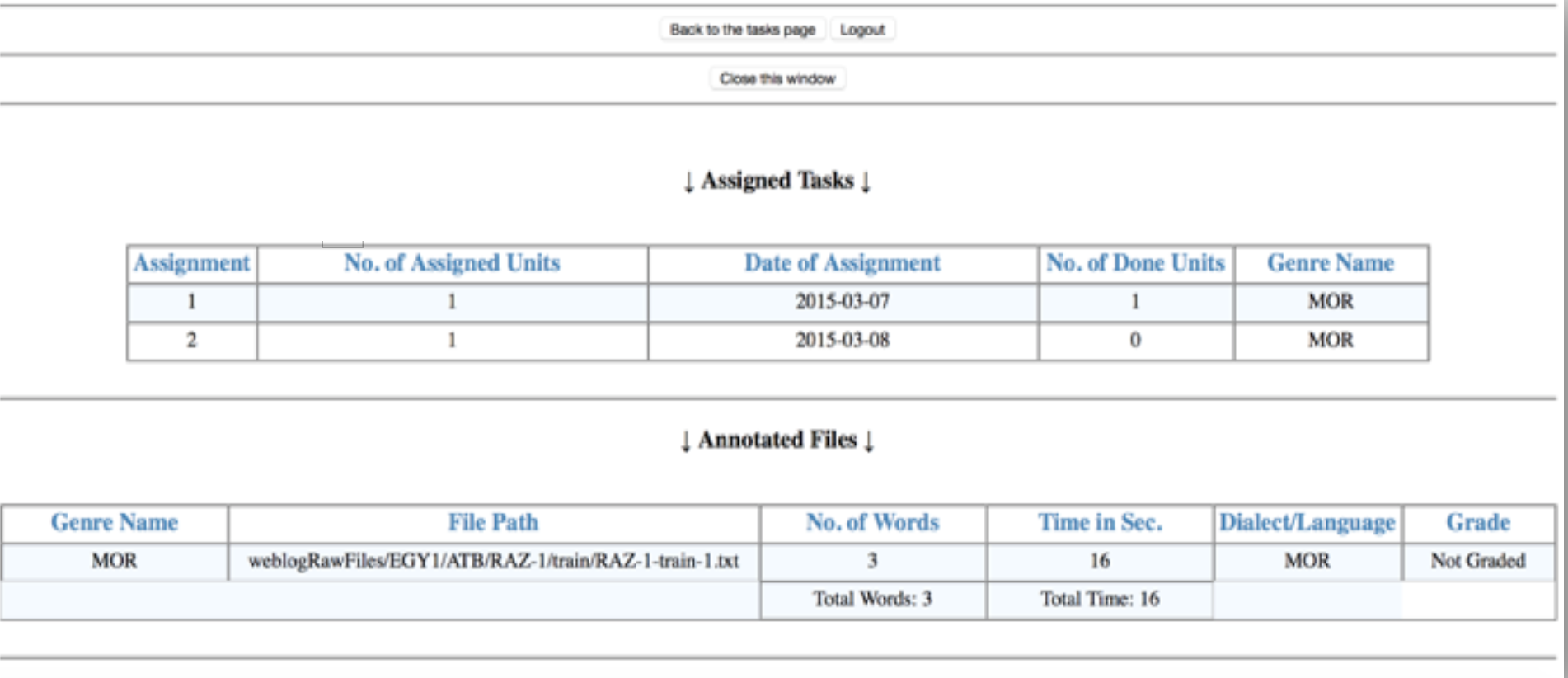}
  \caption{An example of the annotator's "Check-Status" screen }
  \label{fig:picture9.png}
\end{figure*}

\section{Types of Users}

Three types of users have been considered in WASA design: Annotator, Lead-Annotator, and Super-User. Each one of these user types is given and provided with different kinds of permissions, functionalities, and privileges in order to fulfill their tasks.

\subsection{Annotators}
Annotators are provided the following functionalities: \textbf{1-}access assigned tasks; \textbf{2-}annotate the assigned tasks; \textbf{3-} submit annotation; \textbf{4-} check the time needed to submit one unit, e.g., post, or tweets; \textbf{5-} check the grade of the submitted work; \textbf{6-} re-annotate the rejected tasks (by rejected we mean when the annotator received a "No Pass" as a grade on their annotation task); and, \textbf{7-} save work and continue it in a later session.
	
Figure-\ref{fig:screen-shot-1} shows an example of the annotation screen. The words of the posts or tweets that need to be annotated will be displayed as clickable units. When clicked, a pop-up screen appears to allow the annotator to choose the proper tag. To increase the speed of the annotation process, some of the words, like Named Entities and punctuations, will have an initial tag assigned automatically as part of a preprocessing step. However, the annotator is allowed to change the initial tag if he/she finds words annotated with a wrong tag. The interface uses color-coding to reflect useful information and status. For example, 'named entities' will be displayed in purple color, while Other tagged categories such as Latin, URL, punctuation, digits, diacritics, emoticons, sound effects will be displayed in the orange color. Words already annotated will be displayed in blue while words that are yet to be annotated appear in black. Figure-\ref{fig:picture9.png} shows an example of some of the assigned tasks with information about the tasks that have been already submitted (e.g, number of annotated words, speed of annotation, path of the raw file)

\subsection{Lead Annotator} 

For each dialect/language, there is one lead annotator only. Each lead annotator has the following functions: \textbf{1-} Annotator management, e.g., create, edit and delete annotator accounts; \textbf{2-} Tasks management; \textbf{3-} Monitor status and progress; \textbf{4-} Review and grade annotators' work; and \textbf{5-} Produce quality measures like inter-annotator agreement. The system enables lead annotators to reject submitted work that does not meet the assessment criteria and add comments and feedback for the annotators to re-annotate rejected work.

\subsection{Super User} 

There is only one Superuser account in WASA for all dialects/languages. The Superuser functions include: \textbf{1-} Database management and maintenance; \textbf{2-} Lead annotators management, \textbf{3-} Annotators management, \textbf{4-} Monitor the overall performance of the system; and \textbf{5-} Manage annotation data imports and exports.

\section{Data Preprocessing and Input and Output Format}

\begin{figure}[t!]
  \centering
  \small
    \includegraphics[width=0.5\textwidth]{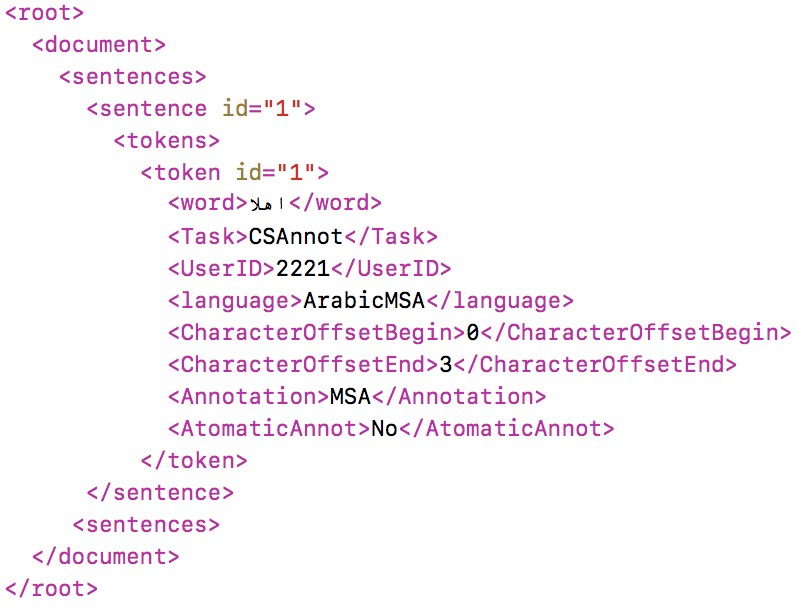}
  \caption{A sample of an output file }
  \label{fig:output}
\end{figure}

\begin{figure}[t!]
  \centering
  \small
    \includegraphics[width=0.5\textwidth]{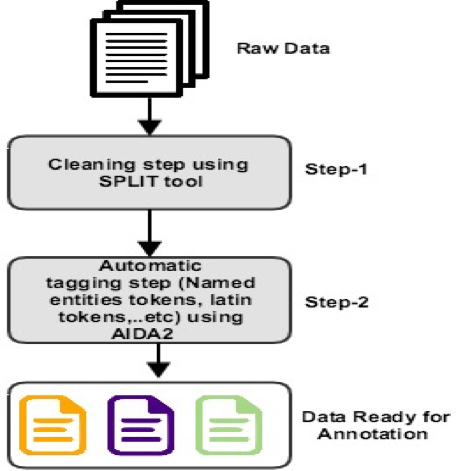}
  \caption{Preprocessing and Cleaning Steps}
  \label{fig:steps}
\end{figure}

The system has the ability to integrate with language-specific date preprocessing scripts to streamline raw data cleaning and preparation. For example, for cleaning process (step-1) the system integrates the Smart Preprocessing (Quasi) Language Independent tool (SPLIT) \cite{AlBadrashiny2016SPLITSP} to handle the encoding issues (i.e, Change the character encoding to UTF8). Moreover, for the Dialectal Arabic (DA) and Modern Standard Arabic (MSA) language pair (step-2), the system integrates with the Automatic Identification of Dialectal Arabic (AIDA2) tool \cite{al2015aida2} to provide initial automatic tagging for named entities (NE), Latin words, URL, punctuation, number, diacritics, emoticon, and speech effects tokens. Figure-\ref{fig:screen-shot-1} illustrates an example of a commentary with some pre-annotated tokens. Named entities tokens are colored purple, while punctuation and numbers are colored with orange.  Both preprocessing and cleaning steps are performed offline. The Super User is the user responsible for preparing the data for annotation. Figure-\ref{fig:steps} shows the cleaning and preprocessing steps.
	The output file is written in a simple XML format as shown in Figure-\ref{fig:output}. The XML file includes all meta-data related to the annotation file such as the annotated, sentence id, task id, language, user id, word id,  actual word, annotation tag, ...etc. The output XML is customizable. The superuser can choose what metadata to be included in the XML output file.  
	
	Our system is able to handle different types of genres such as Twitter, commentaries,  conversations, or discussion forum data. Accordingly,  WASA is quite robust as it is able to handle a variety of data genres and formats.   For example, if the data comes from Twitter, then information like tweet id and user id needs to be preserved along with the annotation tags. If the genre of the data is discussion forums, information such as post order in the context of a conversation thread along with the names of the people who are involved in the conversation are maintained.

\section{Database Design}

WASA system uses a relational database to manage, handle and store all meta-data. The data  stored is categorized as follow:

\subsection{Profiling information}

It saves information about all registered users of the system including their roles (i.e. annotator, lead annotator or superuser), login information as well as the dialect and languages for each one of them. Moreover, It contains information about different languages/dialects used in the project.

\subsection{Annotation Information}

This is the core part of WASA's database. It includes all meta-data related to the annotation tasks such as the number of tasks assigned to each annotator, actual annotations completed by each annotator, and temporarily saved annotations.

\subsection{Assessment Information}
This contains information about \textbf{1)} Task-Annotator assignment: it includes the tasks assigned to each annotator and the number of tasks that have already been annotated and submitted, the number of assigned units (tweets, posts) per each task, genre type, percentage of overlapping units (tweets, posts) shared among annotators to ease the process of calculating inter-annotator agreement, etc.; \textbf{2)} Annotator-Units assignment: It includes information about each unit (post, tweet) that is assigned to the annotators such as post/tweet-id, user-id, genre-id, task-id, path of the assigned file;  Finally \textbf{3)} Language-Unit assignment: It includes information about the language/dialect id for each unit.

\section{Quality Control Measures}

WASA has built-in functionalities that can help in managing the inter-annotator agreement (IAA) measures for different task and report performance statistics. The lead annotator is able to specify the percentage of data annotation overlap between the annotators per task and the system manages to distribute the data and calculate the IAA. Moreover, WASA generates tag distribution, the number of annotated tokens, expected time needed to finish each assigned task, and much other quality management crucial statistics.

\section{Current Status}

We have tested the tool for annotation on Arabic MSA and dialectal data, Chinese-English, Spanish-English, and Hindi-English. The IAA for our the Arabic annotated data is ranged between 92\% and 97\%. Moreover, a small portion of the Code-Switching data that was released in \cite{DiabCS16} was used to test the performance of WASA.  We noticed that the annotators' speed has increased substantially when we assign initial tags to some tags automatically (e.g. punctuation, URL, emoticon, ...etc.). The average time for annotating a full tweet was $\sim$ 40 seconds without using SPLIT tool \cite{AlBadrashiny2016SPLITSP}, but after assigning the initial tags using the SPLIT tool, the average time for annotating a full tweet became $\sim$ 27 seconds. This results in saving much of the effort in annotating these tags.  
        

\section{Conclusion}
We gave a detailed overview of our annotation system WASA. We have shown that WASA allows multiple annotator teams to work on various tasks simultaneously. Also, we have seen that using the SPLIT tool to annotate some specific tokens automatically has helped in saving the effort and time spent by annotators. Moreover, the annotation quality of these tokens is very high.  We will keep updating and modifying the current functionalities of the system as per different users type feedback. Also, we plan to add more functionality that can help in enhancing the speed, quality, and the efficiency of the CS annotation.



%
%
%
%
%

\section{Acknowledgements}

We would like to thank  Mahmoud Ghoneim for his invaluable suggestions and support in the development of WASA. Also, We would like to acknowledge the useful comments by the three anonymous reviewers who helped in making this publication better presented. 

\section{Bibliographical References}
\label{main:ref}

\bibliographystyle{lrec}
\bibliography{xample}

\begin{thebibliography}{}

\bibitem[\protect\citename{Al-Badrashiny \bgroup et al.\egroup
  }2015]{al2015aida2}
Al-Badrashiny, M., Elfardy, H., and Diab, M.~T.
\newblock (2015).
\newblock Aida2: A hybrid approach for token and sentence level dialect
  identification in arabic.
\newblock In {\em CoNLL}, pages 42--51.

\bibitem[\protect\citename{Al-Badrashiny \bgroup et al.\egroup
  }2016]{AlBadrashiny2016SPLITSP}
Al-Badrashiny, M., Pasha, A., Diab, M.~T., Habash, N., Rambow, O., Salloum, W.,
  and Eskander, R.
\newblock (2016).
\newblock Split: Smart preprocessing (quasi) language independent tool.
\newblock In {\em LREC}.

\bibitem[\protect\citename{Aswani and Gaizauskas}2009]{aswani}
Aswani, N. and Gaizauskas, R.
\newblock (2009).
\newblock Evolving a general framework for text alignment: Case studies with
  two south asian languages.
\newblock In {\em Proceedings of the International Conference on Machine
  Translation: Twenty-Five Years On, Cranfield, Bedfordshire, UK, November}.

\bibitem[\protect\citename{Aziz \bgroup et al.\egroup }2012]{aziz2012pet}
Aziz, W., Castilho, S., and Specia, L.
\newblock (2012).
\newblock Pet: a tool for post-editing and assessing machine translation.
\newblock In {\em LREC}, pages 3982--3987.

\bibitem[\protect\citename{Benajiba and Diab}2010]{benajiba2010web}
Benajiba, Y. and Diab, M.
\newblock (2010).
\newblock A web application for dialectal arabic text annotation.
\newblock In {\em Proceedings of the lrec workshop for language resources (lrs)
  and human language technologies (hlt) for semitic languages: Status, updates,
  and prospects}.

\bibitem[\protect\citename{Cunningham \bgroup et al.\egroup }2009]{cunningham}
Cunningham, H., Maynard, D., Bontcheva, K., Tablan, V., Ursu, C., Dimitrov, M.,
  Dowman, M., Aswani, N., Roberts, I., Li, Y., et~al.
\newblock (2009).
\newblock {\em Developing Language Processing Components with Gate Version 5:(A
  User Guide)}.
\newblock University of Sheffield.

\bibitem[\protect\citename{Diab \bgroup et al.\egroup }2010]{diab2010colaba}
Diab, M., Habash, N., Rambow, O., Altantawy, M., and Benajiba, Y.
\newblock (2010).
\newblock Colaba: Arabic dialect annotation and processing.
\newblock In {\em Lrec workshop on semitic language processing}, pages 66--74.

\bibitem[\protect\citename{Diab \bgroup et al.\egroup }2016]{DiabCS16}
Diab, M., Ghoneim, M., Hawwari, A., AlGhamdi, F., AlMarwani, N., and
  Al-Badrashiny, M.
\newblock (2016).
\newblock Creating a large multi-layered representational repository of
  linguistic code switched arabic data.
\newblock In {\em Proceedings of the Tenth International Conference on Language
  Resources and Evaluation (LREC 2016)}, Paris, France, may. European Language
  Resources Association (ELRA).

\bibitem[\protect\citename{Dickinson and Ledbetter}2012]{dickinson}
Dickinson, M. and Ledbetter, S.
\newblock (2012).
\newblock Annotating errors in a hungarian learner corpus.
\newblock In {\em LREC}, pages 1659--1664.

\bibitem[\protect\citename{Kahan \bgroup et al.\egroup }2002]{kahan}
Kahan, J., Koivunen, M.-R., Prud'Hommeaux, E., and Swick, R.~R.
\newblock (2002).
\newblock Annotea: an open rdf infrastructure for shared web annotations.
\newblock {\em Computer Networks}, 39(5):589--608.

\bibitem[\protect\citename{Obeid \bgroup et al.\egroup }2016]{obeid}
Obeid, O., Bouamor, H., Zaghouani, W., Ghoneim, M., Hawwari, A., Alqahtani, S.,
  Diab, M., and Oflazer, K.
\newblock (2016).
\newblock Mandiac: A web-based annotation system for manual arabic
  diacritization.
\newblock In {\em The 2nd Workshop on Arabic Corpora and Processing Tools 2016
  Theme: Social Media}, page~16.

\bibitem[\protect\citename{Samih \bgroup et al.\egroup }2016]{samih2016sawt}
Samih, Y., Maier, W., and Kallmeyer, L.
\newblock (2016).
\newblock Sawt: Sequence annotation web tool.
\newblock {\em EMNLP 2016}, page~65.

\bibitem[\protect\citename{Vargas-Vera \bgroup et al.\egroup }2002]{vargas}
Vargas-Vera, M., Motta, E., Domingue, J., Lanzoni, M., Stutt, A., and
  Ciravegna, F.
\newblock (2002).
\newblock Mnm: Ontology driven semi-automatic and automatic support for
  semantic markup.
\newblock In {\em International Conference on Knowledge Engineering and
  Knowledge Management}, pages 379--391. Springer.

\bibitem[\protect\citename{Yimam \bgroup et al.\egroup }2013]{WebAnnot}
Yimam, S.~M., Gurevych, I., Eckart~de Castilho, R., and Biemann, C.
\newblock (2013).
\newblock Webanno: A flexible, web-based and visually supported system for
  distributed annotations.
\newblock In {\em Proceedings of the 51st Annual Meeting of the Association for
  Computational Linguistics: System Demonstrations}, pages 1--6, Sofia,
  Bulgaria, August. Association for Computational Linguistics.

\end{thebibliography}

\end{document}